\documentclass[11pt,a4paper]{article}
\usepackage[hyperref]{naaclhlt2019}
\usepackage{times}
\usepackage{latexsym}
\usepackage{hyperref}       % hyperlinks
\usepackage{url}            % simple URL typesetting
\usepackage{amsfonts, amssymb}       % blackboard math symbols
\usepackage{nicefrac}       % compact symbols for 1/2, etc.
\usepackage{microtype}      % microtypography
\usepackage{graphicx}
\usepackage{capt-of}
\usepackage{float}
\usepackage{macros}
\usepackage{mathtools}
\usepackage{array}
\usepackage{url}
\usepackage{amsmath}

\usepackage[colorinlistoftodos]{todonotes}

\aclfinalcopy % Uncomment this line for the final submission
\title{Transferable Neural Projection Representations}

\author{Chinnadhurai Sankar \\
  Mila, Universit\'e de Montr\'eal\Thanks{Work done during internship at Google.}\\
  Montreal, QC, Canada\\
  {\tt chinnadhurai@gmail.com} \\\And
  Sujith Ravi  \\
  Google Research \\
  Mountain View, CA, USA  \\
  {\tt sravi@google.com}\\ \And
  Zornitsa Kozareva  \\
  Google \\
   Mountain View, CA, USA \\
  {\tt  zornitsa@kozareva.com} \\}

\date{}

\begin{document}
\maketitle
\begin{abstract}
 Neural word representations are at the core of many state-of-the-art natural language processing models. A widely used approach is to pre-train, store and look up word or character embedding matrices. While useful, such representations occupy huge memory making it hard to deploy on-device and often do not generalize to unknown words due to vocabulary pruning. 
 
 In this paper, we propose a skip-gram based architecture coupled with Locality-Sensitive Hashing (LSH) projections to learn efficient dynamically computable representations. Our model does not need to store lookup tables as representations are computed on-the-fly and require low memory footprint. The representations can be trained in an unsupervised fashion and can be easily transferred to other NLP tasks. For qualitative evaluation, we analyze the nearest neighbors of the word representations and discover semantically similar words even with misspellings. For quantitative evaluation, we plug our transferable projections into a simple LSTM and run it on multiple NLP tasks and show how our transferable projections achieve better performance compared to prior work.
\end{abstract}

\section{Introduction}

Pre-trained word representations are at the core of many neural language understanding models. Among the most popular and widely used word embeddings are word2vec \cite{skipgram_mikolov}, GloVe \cite{glove} and ELMO \cite{elmo}. The biggest challenge with word embedding is that they require lookup and a large memory footprint, as we have to store one entry ($d$-dim vector) per word and it blows up.

In parallel, the tremendous success of deep learning models and the explosion of mobile, IoT devices coupled together with the growing user privacy concerns have led to the need for deploying deep learning models on-device for inference. This has led to new research in compressing large and complex deep learning models for low power on-device deployment. Recently, \citep{sgnn_emnlp18} developed an on-device neural text  classification model. They proposed to reduce the memory footprint of large neural networks by replacing the input word embeddings with projection based representations. \citep{sgnn_emnlp18} used n-gram features to generate binary LSH \cite{Charikar2002} randomized projections on the fly surpassing the need to store word emebdding tables and reducing the memory size. The projection models reduce the memory occupied by the model from $O(|V|)$ to $O(n_{\mathbb{P}})$, where $|V|$ refers to the vocabulary size and $n_{\mathbb{P}}$ refers to number of projection operations \citep{projection_net2017}. Two key advantages of the projection based representations over word embeddings are: (1) they are fixed and have low memory size; (2) they can handle out of vocabulary words. However, the projections in \citep{sgnn_emnlp18} are static and currently do not leverage pre-training on large unsupervised corpora, which is an important property to make the projections transferable to new tasks.

In this paper, we propose to combine the best of both worlds by learning transferable neural projection representations over randomized LSH projections. We do this by introducing new neural architecture inspired by the skip gram model of \cite{skipgram_mikolov} and combined with a deep MLP plugged on top of LSH projections. In order to make this model train better, we introduce new regularizing loss function, which minimizes the cosine similarities of the words within a mini-batch. The loss function is critical for generalization.

In summary, our model (1) requires a fixed and low memory footprint, (2) can handle out of vocabulary words and misspellings, (3) captures semantic and syntactic properties of words; (4) can be easily plugged to other NLP models and (5) can support training with data augmentation by perturbing characters of input words. To validate the performance of our approach, we conduct a qualitative analysis of the nearest neighbours in the learned representation spaces and a quantitative evaluation via similarity, language modeling and NLP tasks.

\section{Neural Projection Model}

We propose a novel model (NP-SG) to learn compact neural representations that combines the benefit of representation learning approaches like skip-gram model with efficient LSH projections that can be computed on-the-fly.

\subsection{Vanilla Skip-Gram Model}

In the skip-gram model \citep{skipgram_mikolov}, we learn continuous distributed representations for words in a large fixed vocabulary, $\mathbb{V}$ to predict the context words surrounding them in documents. We maintain an embedding look up table, $v(w) \in \mathbb{R}^{d}$ for every word, $w \in \mathbb{V}$. 

For each word, $w_{t}$ in the training corpus of size $T$, the set of context words $\mathbb{C}_{t} = \{w_{t-W_{t}},\ldots, w_{t-1},w_{t+1},\ldots,w_{t+W_{t}}\}$ includes $W_{t}$ words to the left and right of $w_{t}$ respectively. $W_{t}$ is the window size randomly sampled from the set $\{1,2,\ldots,N\}$, where $N$ is the maximum window size. Given a pair of words, $\{w_c, w_t\}$, the probability of $w_c$ being within the context window of $w_t$ is given by equation \ref{eq:prob_input_out}.

\begin{equation}
\begin{split}
\label{eq:prob_input_out}
\mathrm{P}(w_c|w_t) & = \sigma(v'(w_c)^\intercal v(w_t)) \\
& = \frac{1}{1 + \mathrm{exp}(-v'(w_c)^\intercal v(w_t))}
\end{split}
\end{equation}
where $v, v'$ are input and context embedding look up tables.

\subsection{Neural Projection Skip-Gram (NP-SG)}

In the neural projection approach, we replace the input embedding look up table, $v(w)$ in equation \ref{eq:prob_input_out} with a deep $n$-layer MLP over the binary projection, $\mathbb{P}(w)$ as shown equation \ref{eq:neuralproj_mlp}.

\begin{equation}
\label{eq:neuralproj_mlp}
v_{\mathbb{P}}(w) = \mathbb{N}(f_n(\mathbb{P}(w)))
\end{equation}
where $v_{\mathbb{P}}(w) \in \mathbb{R}^{d}$, $f_n$ is a $n$-layer deep neural network encoder with $ReLU$ non-linear activations after each layer except for the last layer as shown in Figure \ref{fig:trainable_projection}. $\mathbb{N}$ refers to a normalization applied to the final layer of $f_n$. We experimented with Batch-normalization, L2-normalization and layer normalization; batch-normalization works the best.
{

The binary projection $\mathbb{P}(w)$ is computed using locality-sensitive projection operations~\citep{projection_net2017} which can be performed on-the-fly (i.e., without any embedding look up) to yield a fixed, low-memory footprint binary vector. Unlike~\citep{sgnn_emnlp18} which uses {\it static} projections to encode the entire input text and learn a classifier, NP-SG creates a {\it trainable} deep projection representation for words using LSH projections over character-level features combined with contextual information learned via the skip-gram architecture.

\setlength{\dbltextfloatsep}{-0.5cm}
\setlength{\textfloatsep}{-0.5cm}

\begin{figure}
\centering
    \includegraphics[scale=0.5]{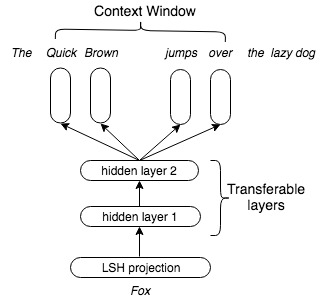}
    \caption{Neural Projection Skip-gram (NP-SG) model}
    \label{fig:trainable_projection}
\end{figure}

\subsection{Training NP-SG Model}

We follow a similar approach as~\citet{skipgram_mikolov} and others  for training our neural projection skip-gram model (NP-SG). We define the training objective to maximize the probability of predicting the context words given the current word. Formally, the model tries to learn the word embeddings by maximizing the objective, $J(\theta)$ known as negative sampling (NEG), given by equation \ref{eq:sg_objective}. 
\begin{equation}
\label{eq:sg_objective}
J(\theta) = \sum_{t=1}^{T} \sum_{w_c \in \mathbb{C}_{t}} J_{w_t,w_c}(\theta)
\end{equation}
%where, 

\begin{equation}
\begin{split}
J_{w_t,w_c}(\theta) & = \mathrm{log}(\mathrm{P}(w_c|w_t)) \\
+ & \sum_{i=1,w_i \sim \mathrm{P}_{n}(w)}^{k}\mathrm{log}(1-\mathrm{P}(w_i|w_t))
\end{split}
\end{equation}

\noindent where $k$ is the number of randomly sampled words from the training corpus according to the noise distribution, $\mathrm{P}_{n}(w) \propto U(w)^{3/4}$, where $U(w)$ is the unigram distribution of the training corpus.

\noindent{\bf Model improvements:} Training an NP-SG model as is, though efficient, may not lead to highly discriminative representations. During training, we noticed that the word representations, $v_{\mathbb{P}}(w)$ were getting projected in a narrow sub-space where the cosine similarities of all the words in the dataset were too close to $1.0$. This made the convergence slower and led to poor generalization.
}

\subsection{Discriminative NP-SG Models}
To encourage the word representations to be more spaced out in terms of the cosine similarities, we introduce an additional explicit regularizing L2-loss function. With the assumption that the words in each mini-batch are randomly sampled, we add a L2-loss
%,$\mathrm{L}_{2}^{cs}$ 
over the cosine similarities between all the words within a mini-batch, as shown in equation \ref{eq:L2_loss}.

\begin{equation}
\label{eq:loss}
Loss = J(\theta) + \mathrm{L}_{2}^{cs}(\textbf{w}_{mb})
\end{equation}

\begin{equation}
\small
\label{eq:L2_loss}
\mathrm{L}_{2}^{cs}(\textbf{w}_{mb}) = \lambda \, \cdot \, \Vert\,\{\mathrm{CS}(w_i,w_j) \, | \, i,j \in [0, mb) \}\,\Vert_{2}^{2}
\end{equation}
where $\mathrm{CS}(w_i,w_j)$ refers to the cosine similarity between $w_i$ and $w_j$, $mb$ refers to the mini-batch size and $\textbf{w}_{mb}$ refers to the words in the mini-batch. 
We enforce this using a simple outerproduct trick. We extract the cosine-similarities between all the words within a mini-batch in a single shot by computing the outer-product of the $L_2$ row normalized word representations corresponding to each minibatch $\hat v_{\mathbb{P}}(\textbf{w}_{mb})$, as shown in equation \ref{eq:outerprodut_trick}. 

\begin{equation}
\small
\label{eq:outerprodut_trick}
\mathrm{L}_{2}^{cs}(\textbf{w}_{mb}) = \frac{\lambda}{2} \, \cdot \, \Vert\, \mathrm{Flatten}(\hat v_{\mathbb{P}}(\textbf{w}_{mb}) \, \cdot \, \hat v_{\mathbb{P}}(\textbf{w}_{mb})^\intercal)\,\Vert_{2}^{2}
\end{equation}

\subsection{Improved NP-SG Training} \label{sec:data_augment}
Since the NP-SG model does not have a fixed vocabulary size, we can be flexible and leverage a lot more information during training compared to standard skip-gram models which require vocabulary pruning for feasibility.

To improve training for NP-SG model, we augment the dataset with inputs words after applying character level perturbations to them. The perturbations are such a way that they are commonly occurring misspellings in documents. We mainly experiment with three types of pertub operation APIs \citep{char_perturbation}.

\begin{itemize}
\item \textit{insert(word, n)} : We randomly choose \textit{n} chars from the character vocabulary and insert them randomly into the input \textit{word}. We ignore the locations of first and last character in the word for the \textit{insert} operation. Example transformation: $sample \rightarrow samnple$.

\item \textit{swap(word, n)} : We randomly swap the location of two characters in the word \textit{n} times. As with the \textit{insert} operation, we ignore the first and last character in the word for the \textit{swap} operation. Example transformation: $sample \rightarrow sapmle$.

\item \textit{duplicate(word, n)} : We randomly duplicate a character in the word by \textit{n} times. Example transformation: $sample \rightarrow saample$.
\end{itemize}

We would like to note that the perturbation operations listed above are not exhaustive and we plan to experiment with more operations in the future.

%\section{Skip-Gram Training Setup}
\section{Training Setup}
\subsection{Dataset}
We train our skipgram models on the wikipedia data XML dump, \textit{\textbf{enwik9}}\footnote{http://mattmahoney.net/dc/enwik9.zip}. We extract the normalized English text  from the XML dump using the Matt Mahoney’s pre-processing perl script\footnote{http://mattmahoney.net/dc/textdata}. 

We fix the vocabulary to the top $100 k$ frequently occurring words.
We sub-sample words in the training corpus, dropping them with probability, $\mathrm{P}(w) = 1 - \sqrt{t/freq(w)}$, where $freq(w)$ is the frequency of occurrence of $w$ in the corpus and we set the threshold, $t$ to $10^{-5}$. We perturb the input words with a probability of $0.4$ using a randomly chosen perturbation described in Section \ref{sec:data_augment}.

\begin{table*}[ht!]
\centering
%%%%\begin{tabular}{ m{4.5em}  m{2.5em} m{2.5em} m{2.5em} m{2.5em} m{2em}} 

\begin{tabular}{c|  c|  c | c|c| c} %@{ }c@{ }|
 \hline
\textit{Dataset} & \textit{SG (10M)} & \textit{NP-SG} (w/oOP) & \textit{NP-SG} (1M) &\textit{NP-SG} (2M) & \textit{NP-SG} (4M)\\
 \hline

EN-MTurk-287 & 0.5409 & 0.0107 & \bf 0.5629 & \bf 0.5517 & \bf 0.5494\\
EN-WS-353-ALL & 0.5930 & 0.0710 & 0.4891 & 0.5215 & 0.5370\\
EN-WS-353-REL & 0.5359 & 0.0203 & 0.4956 & \bf 0.5746 &\bf 0.5671\\
EN-WS-353-SIM & 0.6242 & 0.1043 & 0.4994 & 0.5116  & 0.5111\\ 
EN-RW-STANFORD & 0.1505 & 0.0401 & 0.0184 & 0.0375 & 0.0835\\
EN-VERB-143 & 0.2452 & 0.0730 & 0.1333 & 0.1500 & 0.2108 \\
 \hline
\end{tabular}
\caption{Similarity Tasks: \# of params, $100k$ vocabulary size for skipgram baseline, $100$ embedding size.} \label{table:similarity_tasks}
\end{table*}

\subsection{Implementation Details}
We fix the number of random projections to $80$ and the projection dimension to $14$. We use a 2-layer MLP~(sizes: $[2048, 100]$) regularized with dropout (with probability of $0.65$) and weight decay (regularization parameter of $0.0005$) to transform the binary random projections to continuous word representation. 
For the vanilla skipgram model, we fix the embedding size to $100$.
For both models, we use $25$ negative samples for the NEG loss. We learn the parameters using the Adam optimizer \cite{kingma2014adam} with a default learning rate of $0.001$, clipping the gradients which have a norm larger than $5.0$.
We initialize the weights of the MLP using Xavier initialization, and output embeddings uniformly random in the range $[-1.0, 1.0]$. We use a batch size of $1024$ in all our experiments. We found that $\lambda=0.01$ for the outerproduct loss to be working better after experimenting with other values. Training time for our model was around $0.85$ times that of the skipgram model. Both the models were trained for $10$ epochs.

\section{Experiments}

We show both qualitative and quantitative evaluation on multiple tasks for the NP-SG model.

\subsection{Qualitative Evaluation and Results}

Table \ref{table:nn} shows the nearest neighbors produced by NP-SG for select words. Independent of whether it is an original or misspelled word, our NP-SG model accurately retrieves relevant and semantically similar words.

\begin{table}[h!]
\small
\begin{tabular}{ l| l} 
%%%\begin{tabular}{ m{3em} m{6cm}} 

\hline
\it Word & \it Nearest neighbours\\
 \hline 
king & reign, throne, kings, knights, vii, regent \\
kingg & vii, younger, peerage, iv, tiberius, frederick  \\
\hline
woman & man, young, girl, child, girls, women \\
wwoamn & man, herself, men, couple, herself, alive \\
\hline

city & town, village, borough, township, county \\
ciity & town, village, borough, county, unorganized\\
\hline

time & few, times, once, entire, prominence, since\\
tinme & times, once, takes, taken, another, only\\

\hline
zero & two, three, seven, one, eight, four\\
zzero & two, three, five, six, seven, four\\

 \hline
\end{tabular}
 \caption{Sampled nearest neighbors for NP-SG.}\label{table:nn}

\end{table}

\subsection{Quantitative Evaluation and Results}

We evaluate our NP-SG model on similarity, language modeling and text classification tasks. Similarity tests the ability to capture words, while language modeling and classification warrant the ability to transfer the neural projections.

\subsubsection{Similarity Task}

We evaluate our NP-SG word representations on 4 different widely used benchmark datasets for measuring similarities. 

\noindent{\bf Dataset:} \textit{MTurk-287} \citep{EN-Mturk-287} has 287 pairs of words and was constructed by crowdsourcing the human similarity ratings using Amazon Mechanical Turk. \textit{WS353} \citep{WS353} has 353 pairs of similar English words rated by humans and is further split into \textit{WS353-SIM}. \textit{WS353-REL} \citep{ws353_rel_sim} captures different types of similarities and relatedness. \textit{RW-STANFORD} \citep{stanford_rare_word} has 2034 rare word pairs sampled from different frequency bins.

\noindent{\bf Evaluation:} For all the datasets, we compute the Spearman’s rank correlation coefficient between the rankings computed by skip-gram models (baseline SG and NP-SG) and the human rankings.
We use cosine similarity metric to measure word similarity.

\noindent{\bf Results:} Table \ref{table:similarity_tasks} shows that NP-SG, with significantly smaller number of parameters comes close to the skip-gram model (SG) and even outperforms it with 2.5x-10x compression. NP-SG gets better representations even with misspellings which cannot be handled by vanilla SG.

It is interesting to note that the vanilla skip-gram model does well on \textit{WS353-SIM} compared to \textit{WS353-REL}. This behavior is reversed in our NP-SG model, which indicates that it captures meronym-holonym relationships better than the vanilla skip-gram model. Although NP-SG handles out of vocabulary words in the form of misspellings, it needs further improvement for rare word similarity task. We plan to improve it by including  context word n-gram features in the LSH projection function, allowing NP-SG to also  leverage information from the context words in the case of rare words and provide word sense disambiguation. 

\subsubsection{Language Modeling}

We applied NP-SG to language modeling task on the Penn Treebank (PTB)\citep{ptb} dataset. 
We consider a single layer LSTM with hidden size of 2048 for the language model task. 
With the input embedding size of 200, we observed a perplexity of $\approx 120$ on the test set after training for $5$ epochs.
We replace the input embeddings in the LSTM with transferable encoder layer of the NP-SG model. 
We train the LSTMs with and without pretrained initializations. 
Since we observed convergence issues with the single layer NP-SG representation, we considered 2-layer MLP with layer sizes (1024, 256) for the NP-SG representations. 
We found that while the model without pretrained NP-SG layer got stuck at a perplexity of around $300$, the pretrained version converged to a perplexity of $140$, comparable to the embedding based network. 
We leave the analysis of the impact of the deeper NP-SG layers to the future work.

\subsubsection{Text Classification}

For the text classification evaluations, we used two different tasks and datasets. For the dialog act classification task, we used the MRDA dataset from the ICSI Meeting Recorder Dialog Act Corpus \cite{Janin03theicsi}. MRDA is a multiparty dialog annotated with 5 dialog act tags. For the question classification task, we used the TREC dataset \cite{TREC}. The task is given a question to predict the most relevant category.

We trained a single layer LSTM (hidden size: $256$) with and without the pretrained NP-SG layers. Overall, we observed accuracy improvements of $+5.7\%$ and $+3.75\%$ compared to  baseline models without pretrained NP-SG initializations on TREC and MRDA respectively.

\section{Conclusion}
In this paper, we introduced a new neural architecture (NP-SG), which learns transferable word representations that can be efficiently and dynamically computed on device without any embedding look up. We proposed an unsupervised method to train the new architecture and learn more discriminative word representations. We compared the new model with a skip-gram approach and showed qualitative and quantitative comparisons on multiple language tasks. The evaluations show that our NP-SG model learns better representations even with misspellings and reaches competitive results with skip-gram on similarity tasks, even outperforming with 2.5x-10x fewer parameters.

\section*{Acknowledgments}
The authors would like to thank the Google Expander team for many helpful discussions.

\bibliography{naaclhlt2019}

\begin{thebibliography}{15}
\expandafter\ifx\csname natexlab\endcsname\relax\def\natexlab#1{#1}\fi

\bibitem[{Adam et~al.(2003)Adam, Baron, Edwards, Ellis, Gelbart, Morgan,
  Peskin, Pfau, Shriberg, Stolcke, and Wooters}]{Janin03theicsi}
Janin Adam, Don Baron, Jane Edwards, Dan Ellis, David Gelbart, Nelson Morgan,
  Barbara Peskin, Thilo Pfau, Elizabeth Shriberg, Andreas Stolcke, and Chuck
  Wooters. 2003.
\newblock The icsi meeting corpus.
\newblock In \emph{Proceedings of the 5TH SIGdial Workshop on Discourse and
  Dialogue}, pages 364--367.

\bibitem[{Agirre et~al.(2009)Agirre, Alfonseca, Hall, Kravalova, Pasca, and
  Soroa}]{ws353_rel_sim}
Eneko Agirre, Enrique Alfonseca, Keith~B. Hall, Jana Kravalova, Marius Pasca,
  and Aitor Soroa. 2009.
\newblock A study on similarity and relatedness using distributional and
  wordnet-based approaches.
\newblock In \emph{Human Language Technologies: Conference of the North
  American Chapter of the Association of Computational Linguistics,
  Proceedings, May 31 - June 5, 2009, Boulder, Colorado, {USA}}, pages 19--27.

\bibitem[{Charikar(2002)}]{Charikar2002}
Moses~S. Charikar. 2002.
\newblock \href {https://doi.org/10.1145/509907.509965} {Similarity estimation
  techniques from rounding algorithms}.
\newblock In \emph{Proceedings of the Thiry-fourth Annual ACM Symposium on
  Theory of Computing}, STOC '02, pages 380--388, New York, NY, USA. ACM.

\bibitem[{Finkelstein et~al.(2001)Finkelstein, Gabrilovich, Matias, Rivlin,
  Solan, Wolfman, and Ruppin}]{WS353}
Lev Finkelstein, Evgeniy Gabrilovich, Yossi Matias, Ehud Rivlin, Zach Solan,
  Gadi Wolfman, and Eytan Ruppin. 2001.
\newblock Placing search in context: the concept revisited.
\newblock In \emph{Proceedings of the Tenth International World Wide Web
  Conference, {WWW} 10, Hong Kong, China, May 1-5, 2001}, pages 406--414.

\bibitem[{Gao et~al.(2018)Gao, Lanchantin, Soffa, and Qi}]{char_perturbation}
Ji~Gao, Jack Lanchantin, Mary~Lou Soffa, and Yanjun Qi. 2018.
\newblock Black-box generation of adversarial text sequences to evade deep
  learning classifiers.
\newblock In \emph{2018 {IEEE} Security and Privacy Workshops, {SP} Workshops
  2018, San Francisco, CA, USA, May 24, 2018}, pages 50--56.

\bibitem[{Kingma and Ba(2014)}]{kingma2014adam}
Diederik Kingma and Jimmy Ba. 2014.
\newblock Adam: A method for stochastic optimization.
\newblock \emph{arXiv preprint arXiv:1412.6980}.

\bibitem[{Lin and Katz(2006)}]{TREC}
Jimmy~J. Lin and Boris Katz. 2006.
\newblock Building a reusable test collection for question answering.
\newblock \emph{{JASIST}}, 57(7):851--861.

\bibitem[{Luong et~al.(2013)Luong, Socher, and Manning}]{stanford_rare_word}
Thang Luong, Richard Socher, and Christopher~D. Manning. 2013.
\newblock Better word representations with recursive neural networks for
  morphology.
\newblock In \emph{Proceedings of the Seventeenth Conference on Computational
  Natural Language Learning, CoNLL 2013, Sofia, Bulgaria, August 8-9, 2013},
  pages 104--113.

\bibitem[{Mikolov et~al.(2013)Mikolov, Sutskever, Chen, Corrado, and
  Dean}]{skipgram_mikolov}
Tomas Mikolov, Ilya Sutskever, Kai Chen, Gregory~S. Corrado, and Jeffrey Dean.
  2013.
\newblock Distributed representations of words and phrases and their
  compositionality.
\newblock In \emph{Advances in Neural Information Processing Systems 26: 27th
  Annual Conference on Neural Information Processing Systems 2013. Proceedings
  of a meeting held December 5-8, 2013, Lake Tahoe, Nevada, United States.},
  pages 3111--3119.

\bibitem[{Pennington et~al.(2014)Pennington, Socher, and Manning}]{glove}
Jeffrey Pennington, Richard Socher, and Christopher Manning. 2014.
\newblock Glove: Global vectors for word representation.
\newblock In \emph{Proceedings of the 2014 Conference on Empirical Methods in
  Natural Language Processing (EMNLP)}, pages 1532--1543. Association for
  Computational Linguistics.

\bibitem[{Peters et~al.(2018)Peters, Neumann, Iyyer, Gardner, Clark, Lee, and
  Zettlemoyer}]{elmo}
Matthew Peters, Mark Neumann, Mohit Iyyer, Matt Gardner, Christopher Clark,
  Kenton Lee, and Luke Zettlemoyer. 2018.
\newblock Deep contextualized word representations.
\newblock In \emph{Proceedings of the 2018 Conference of the North American
  Chapter of the Association for Computational Linguistics: Human Language
  Technologies, Volume 1 (Long Papers)}, pages 2227--2237. Association for
  Computational Linguistics.

\bibitem[{Radinsky et~al.(2011)Radinsky, Agichtein, Gabrilovich, and
  Markovitch}]{EN-Mturk-287}
Kira Radinsky, Eugene Agichtein, Evgeniy Gabrilovich, and Shaul Markovitch.
  2011.
\newblock A word at a time: computing word relatedness using temporal semantic
  analysis.
\newblock In \emph{Proceedings of the 20th International Conference on World
  Wide Web, {WWW} 2011, Hyderabad, India, March 28 - April 1, 2011}, pages
  337--346.

\bibitem[{Ravi(2017)}]{projection_net2017}
Sujith Ravi. 2017.
\newblock Projectionnet: Learning efficient on-device deep networks using
  neural projections.
\newblock \emph{CoRR}, abs/1708.00630.

\bibitem[{Ravi and Kozareva(2018)}]{sgnn_emnlp18}
Sujith Ravi and Zornitsa Kozareva. 2018.
\newblock Self-governing neural networks for on-device short text
  classification.
\newblock In \emph{Proceedings of the 2018 Conference on Empirical Methods in
  Natural Language Processing, Brussels, Belgium, October 31 - November 4,
  2018}, pages 804--810.

\bibitem[{Taylor et~al.(2003)Taylor, Marcus, and Santorini}]{ptb}
Ann Taylor, Mitchell Marcus, and Beatrice Santorini. 2003.
\newblock The penn treebank: An overview.

\end{thebibliography}
\bibliographystyle{acl_natbib}
\end{document}